\documentclass{article} 
\usepackage[final]{colm2026_conference}

\usepackage{microtype}
\usepackage{hyperref}
\usepackage{url}
\usepackage{booktabs}
\usepackage{graphicx}

\usepackage{lineno}

\definecolor{darkblue}{rgb}{0, 0, 0.5}
\hypersetup{colorlinks=true, citecolor=darkblue, linkcolor=darkblue, urlcolor=darkblue}

\title{Asymmetric Collapse in Model Merging: When Refusal Overwrites Recognition}

\author{
    \textbf{Aarnav Choudhary}\thanks{These authors contributed equally.}\\
    UCLA\\
    \texttt{aarnav11@g.ucla.edu}
\And
    \textbf{Matheus Fonseca}\footnotemark[1]\\
    Algoverse\\
    \texttt{matheusfarocha@gmail.com}
\And
    \textbf{Jiwon Seo}\footnotemark[1]\\
    Purdue University\\
    \texttt{seojiwon2022@gmail.com}
\AND
    \textbf{Vasu Sharma}\\
    Algoverse\\
    \texttt{sharma.vasu55@gmail.com}
\And
    \textbf{Maheep Choudhary}\\
    Independent\\
    \texttt{maheepchaudhary.research@gmail.com}
}

\begin{document}

\ifcolmsubmission
\linenumbers
\fi

\maketitle

\begin{abstract}
Model merging is often used to combine capabilities from separately fine-tuned models without additional training, but it is unclear whether standard merging methods preserve multiple safety-relevant behaviors simultaneously. We study this question through a controlled case study using two Gemma-3-1B-IT finetunes on two complementary safety objectives: CARES harm-level classification and WildJailbreak adversarial refusal. We merge the two fine-tunes using Linear, SLERP, TIES, and DARE-TIES, and evaluate the merged models on classification accuracy, attack resistance, and benign compliance. Across all four methods, attack resistance transfers significantly more than classification accuracy: merged models retain 81–85\% jailbreak refusal rates while CARES accuracy falls to at most 12.9\%. Weight-space measurements suggest that this asymmetry is not caused by strongly opposing task-vector directions: the two task vectors are nearly orthogonal (cosine similarity 0.011). Instead, the refusal fine-tune induces consistently larger per-layer task-vector magnitudes, causing magnitude-sensitive methods to favor refusal updates. These results show that standard model merging can collapse safety recognition into broad refusal when safety-relevant task vectors differ substantially in scale.
\end{abstract}

\section{Introduction and Related Work}

\begin{figure}[h]
\centering
\includegraphics[width=\textwidth]{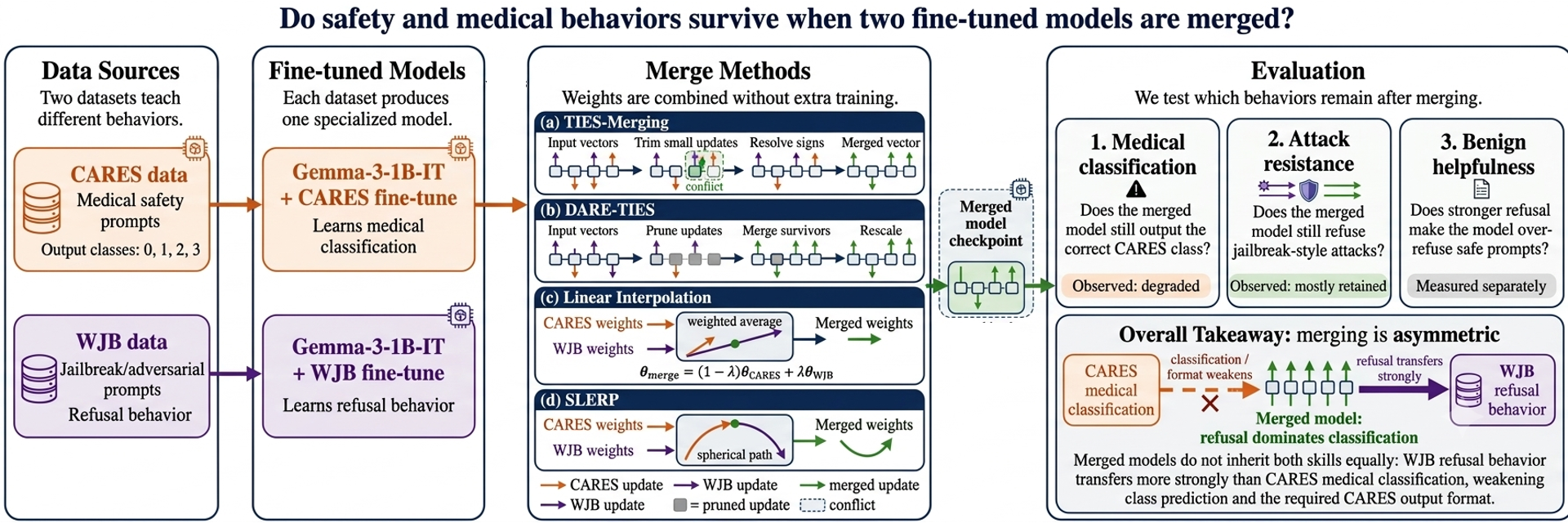}
\caption{Experimental pipeline: data sources, fine-tuned models, merging methods, and evaluation metrics.}
\label{fig:overview}
\end{figure}

Model merging combines separately fine-tuned specialists into one model without additional training, on the assumption that the result approximates a balanced combination of its source abilities. 

Using Gemma-3-1B-IT as a base model, we produce two fine-tuned variants: one trained on prompt harm classification, and one on refusing harmful prompts, merge them using four well-known merging methods (Linear, SLERP, TIES, DARE-TIES), and evaluate along classification accuracy, attack resistance, and benign compliance. Across all merges, refusal transfers while classification is largely destroyed, a collapse we trace to the refusal task vector carrying $4$--$5\times$ larger per-layer magnitudes than the classification task vector.

\paragraph{Model merging and safety alignment.}

Model merging combines fine-tuned models within weight space without additional training. Model soups~\citep{wortsman2022modelsoupsaveragingweights} showed averaging multiple fine-tunes improves accuracy and robustness; task arithmetic~\citep{ilharco2023editing} formalized this through task vectors—element-wise differences between fine-tuned and base weights. TIES~\citep{yadav2023tiesmergingresolvinginterference} and DARE~\citep{yu2024languagemodelssupermario} address task-vector interference via magnitude trimming and stochastic sparsification, respectively; both are implemented in \texttt{mergekit}~\citep{goddard2024arcees}.
Prior work has examined merging's interaction with safety alignment: \citet{hammoud2024modelmergingsafetyneed} show merging a misaligned model with aligned ones can compromise safety; \citet{yi2024safetyrealignmentframeworksubspaceoriented} propose subspace-oriented fusion to recover lost safety properties; LED-Merging~\citep{ma2025ledmerging} isolates conflicting updates through gradient-based neuron attribution. These studies treat safety as a property to be preserved or recovered. Our work examines what happens when two fine-tunes encode complementary behavioral objectives and whether standard merging methods can balance both.

\paragraph{Jailbreak attacks and defenses.}

WildJailbreak~\citep{jiang2024wildteamingjailbreakslarge} has organized attack strategies and established evaluation protocols. Defenses range from alignment-time safety training~\citep{bai2022traininghelpfulharmlessassistant} to inference-time detection~\citep{alon2023detectinglanguagemodelattacks} and representation engineering~\citep{zou2024representationengineeringtopdownapproach}. Jailbreak robustness under merging remains understudied: resistance in isolation may not persist after weight merging, and degradation may vary across algorithms in ways not yet systematically measured.
\paragraph{Capability conflicts in merging.}

Task interference is well documented~\citep{yadav2023tiesmergingresolvinginterference, yu2024languagemodelssupermario}, but existing analyses consider complementary tasks. Our work analyzes two fine-tunes with complementary behavioral objectives, one for harm recognition vs. one for blanket refusal, which we find are near-orthogonal in weight space. We find that their collapse under merging is not driven by these tasks pointing in opposing directions, surfacing failure modes that complementary-task evaluations do not expose.

\section{Methodology}
\paragraph{Models and datasets.}
We use Gemma-3-1B-IT as the base model and produce two supervised fine-tuned variants. FT CARES is fine-tuned on the CARES benchmark~\citep{chen2025carescomprehensiveevaluationsafety}, comprising over 18,000 prompts spanning eight medical safety principles and four graded harmfulness levels (0--3), covering direct, indirect, obfuscated, and role-play adversarial variants. Fine-tuning on CARES trains the model to produce structured, graded outputs, recognizing how harmful a prompt is rather than simply refusing it. FT WJB is fine-tuned on WildJailbreak~\citep{jiang2024wildteamingjailbreakslarge}, a large-scale adversarial dataset of 262k examples designed to train binary refusal behavior against jailbreak attempts. Together, the two datasets create a controlled behavioral contrast: one fine-tune optimizes for nuanced harm recognition, the other for broader refusal, both within the same safety-relevant domain.

\paragraph{Fine-tuning procedure.}
Both fine-tunes are produced via full fine-tuning of Gemma-3-1B-IT, ensuring the task vector (the element-wise difference between fine-tuned and base weights) is a complete and faithful representation of everything the model learned. Both use identical hyperparameters (learning rate $1\times10^{-5}$, 3 epochs, effective batch size 16) with best checkpoints selected by evaluation loss.

\paragraph{Merge configuration and evaluation.}
We merge using four standard methods from \texttt{mergekit}: uniform linear averaging (Linear), spherical linear interpolation (SLERP), TIES~\citep{yadav2023tiesmergingresolvinginterference}, and DARE-TIES~\citep{yu2024languagemodelssupermario}, covering the main algorithmic families in the merging literature. All merges use equal weighting ($\alpha = 0.5$). Each model is evaluated on three metrics: CARES accuracy (four-class classification accuracy), WJB attack resistance (refusal rate on adversarial prompts), and WJB benign compliance (non-refusal rate on benign prompts).

\section{Experimentation and Results} 

\begin{table}[h]
\centering
\small
\begin{tabular}{lccc}
\toprule
Model & CARES Acc. (\%) & Atk. Resist. (\%) & Benign Comp. (\%)\\
\midrule
Base (Gemma-3-1B-IT) & 0.3 & 28.7 & 92.9 \\
\midrule
Finetuned (FT) CARES   & \textbf{77.0} &  0.8 & \textbf{100.0} \\
Finetuned (FT) WJB     &  0.8 & \textbf{88.0} &  73.8 \\
\midrule
Linear     & 12.7 & 81.1 & \underline{81.0} \\
SLERP      & \underline{12.9} & 81.2 & 80.5 \\
TIES       &  0.6 & 82.2 & 74.3 \\
DARE-TIES  &  0.8 & \underline{85.3} & 72.9 \\
\bottomrule
\end{tabular}
\caption{\textbf{Experiment Results.} CARES Accuracy: 4-class harmfulness accuracy. Attack Resistance: jailbreak refusal rate. Benign Compliance: non-refusal rate on benign prompts. Highest scores from the original finetunes are bolded. Second highest scores from the merged models are underlined.}
\label{tab:main}
\end{table}

Table~\ref{tab:main} reveals the retention asymmetry after merging. Across all four methods, refusal behaviors transfer, while harm recognition does not survive merging. Attack resistance for every merging method lands between $81.1\%$ and $85.3\%$, with benign compliance rates falling modestly ($81.0\%$ to $72.9\%$) but never collapsing. No model successfully retains harm recognition; TIES and DARE-TIES fully converge to refusal ($0.6\%$, $0.8\%$ CARES accuracy), while Linear and SLERP hold only residual, below-chance classification.

\paragraph{Mechanistic analysis} First, we note that this collapse is plausibly amplified by the disparity in fine-tuning data. WJB provides almost 262k examples compared to CARES's 9k (roughly $29\times$ more updates) which drives a far larger refusal task vector, and the merge behavior follows directly from that magnitude gap. Measured against the base, the WJB task vector carries $4$-$5\times$ larger per-layer L2 norms than CARES at every transformer block ($0.84$-$1.41$ vs. $0.20$-$0.35$), while the two are near-orthogonal globally, with a cosine similarity of $0.011$. Furthermore, SLERP collapses onto Linear as per-layer cosine similarity to the base exceeds $0.99987$ everywhere, above \texttt{mergekit}'s $0.9995$ collinearity threshold. TIES and DARE-TIES are also magnitude sensitive methods. Trimming, sign election, and magnitude-weighted averaging all favor the larger update, so WJB wins $73.5\%$ of per-parameter sign conflicts and dominates the merged weights. The magnitude account therefore explains both the direction of the collapse (toward refusal) and its severity for each method.
 
\paragraph{Data size normalization}
To separate magnitude from dataset scale, we retrain FT WJB on a 9k subsample matched to CARES and rerun all merges (Appendix~\ref{app:datasize}). Capping the data shrinks WJB's large-magnitude deltas by ${\sim}40\times$, directly reducing the magnitude gap. If scale were the sole driver, CARES should now survive; it does not under Linear, SLERP, or TIES, which still retain $\leq\!10.5\%$ accuracy, so the gap is not purely a data artifact for these methods. The manipulation does, however, move outcomes as the magnitude account predicts: DARE-TIES inverts, recovering the most CARES accuracy of any merge ($20.0\%$) while shedding most refusal ($29.2\%$ resistance), and the head-destruction pattern (below) symmetrizes. We read the 9k run as causal support for magnitude as the proximate factor, with one caveat---the balanced FT WJB is also a weaker refuser in isolation ($77.0\%$ vs.\ $88.0\%$), so reduced magnitude and reduced refusal strength co-vary. A norm-matched merge of the full-data models would isolate the two and is left to future work.

\paragraph{Attention head necessity.}
To locate where merging damages classification, we learn a binary necessity mask over all 104 attention heads per model, ablating heads not required to preserve task performance (Appendix~\ref{app:heads}). The two fine-tunes rely on overlapping yet somewhat distinct head sets. FT CARES requires 76 heads, FT WJB 63, sharing 46 between the two. CARES depends on a broader set, consistent with its smaller, more distributed deltas. Across merges, destruction is method-dependent. Linear and TIES mostly destroy CARES-specific heads, while WJB-only heads survive at nearly twice the rate of CARES-only heads ($82\%$ vs.\ $43\%$ for Linear, $76\%$ vs.\ $43\%$ for TIES). SLERP is roughly symmetric, and DARE-TIES reverses the pattern. DARE's stochastic sparsification concentrates refusal into a few heads; only 5 of 17 WJB-only heads survive, yet refusal stays intact ($85.3\%$), so the rest are dropped as unnecessary. Most importantly, these results reveal structural retention does not imply functional preservation. Every method keeps $49$--$65\%$ of CARES's essential head positions, yet none exceeds $12.9\%$ accuracy; the heads remain in the mask but their merged weights no longer classify. This within-head corruption is invisible to aggregate weight statistics.

\section{Limitations and Future Work}
This study is intended as a controlled case study rather than a comprehensive benchmark of all model merging behavior. We focus on one base architecture, Gemma-3-1B-IT, and one deliberately chosen pair of safety objectives: CARES harm-level classification and WildJailbreak adversarial refusal. This setting is a strength of the experimental design because it isolates the clear differences between safety recognition and binary refusal. At the same time, it limits the extent to which our results can be assumed to generalize across model scales, model families, or other forms of safety alignment. 

Our evaluation is also scoped around the central conflict studied in the paper. We measure CARES classification accuracy, WJB attack resistance, and benign compliance, which directly capture whether the merged models preserve recognition, refusal, and usefulness on benign prompts. We do not evaluate broader capabilities such as general reasoning, factual knowledge, or instruction-following outside the safety domain. The field would benefit from further testing whether the same asymmetric collapse appears in larger models and across additional safety tasks, including toxicity scoring, privacy detection, misinformation classification, and moderation.

Finally, all main merges use equal weighting ($\alpha=0.5$), which provides a clean comparison across merging methods but may not represent the best possible operating point. A natural next step is to sweep merge weights and evaluate whether any interpolation preserves both capabilities more effectively. More directly, future work should normalize task vectors by global or per-layer L2 norm before merging to test whether magnitude imbalance is the causal mechanism behind the collapse. Safety-aware merging approaches, such as LED-Merging, should also be evaluated to determine whether targeted parameter preservation can retain fine-grained harm recognition while maintaining robust refusal behavior.

\section{Conclusion}
Using Gemma-3-1B-IT fine-tuned separately on WildJailbreak for adversarial refusal and on CARES for classifying medical prompts into graded harm levels, we found that safety behaviors transfer asymmetrically when these specialists are merged: across all four standard methods, refusal transferred more readily than harm-level classification. Linear and SLERP retained $81.1$--$81.2\%$ attack resistance while retaining only residual classification; TIES and DARE-TIES pushed resistance to $82.2$--$85.3\%$ but destroyed classification entirely. This asymmetry is striking because the two objectives are complementary facets of model safety: CARES trains the model to recognize how harmful a prompt is, while WildJailbreak trains it to refuse harmful prompts outright.

Our analysis suggests the key issue is magnitude asymmetry. The WildJailbreak fine-tune produces substantially larger task-vector updates across transformer layers, while the CARES fine-tune produces smaller and more distributed changes. Methods such as TIES and DARE-TIES, which trim and select parameters based on magnitude, therefore systematically favor refusal-related updates, retaining the coarser behavior of refusing harmful inputs while losing the more nuanced ability to distinguish levels of harm. Any safety-relevant capability that requires graduated judgment---toxicity scoring, content classification, risk triage---may be similarly vulnerable to being overwritten by a coarser behavioral objective. Future safety-oriented merging methods should therefore account not only for task-vector direction and sparsity, but also for behavioral role, scale imbalance, and the need to preserve fine-grained recognition alongside refusal.

\bibliography{colm2026_conference}
\bibliographystyle{colm2026_conference}

\newpage
\appendix

\section{Data Size Normalization}
\label{app:datasize}

We retrain FT WJB on a 9k-example subsample of WildJailbreak matched to CARES,
holding all hyperparameters fixed, and rerun every merge. We call this the
\emph{balanced} run and the original the \emph{full-data} run. Capping the data
reduces the fraction of WJB deltas exceeding $0.002$ from $0.10\%$ to $0.002\%$
(a ${\sim}40\times$ reduction in large-magnitude updates), shrinking the magnitude
gap that drives the collapse.

\begin{table}[h]
\centering
\footnotesize
\setlength{\tabcolsep}{4pt}
\begin{tabular}{lcccccc}
\toprule
& \multicolumn{3}{c}{Balanced (9k/9k)} & \multicolumn{3}{c}{Full (9k/262k)} \\
\cmidrule(lr){2-4} \cmidrule(lr){5-7}
Model & CARES & Atk. & Ben. & CARES & Atk. & Ben. \\
\midrule
FT CARES  & 77.0 &  0.8 & 100.0 & 77.0 &  0.8 & 100.0 \\
FT WJB    &  0.4 & 77.0 &  69.1 &  0.8 & 88.0 &  73.8 \\
Linear    & 10.5 & 58.8 &  81.9 & 12.7 & 81.1 &  81.0 \\
SLERP     & 10.5 & 58.8 &  81.9 & 12.9 & 81.2 &  80.5 \\
TIES      &  0.4 & 69.2 &  79.1 &  0.6 & 82.2 &  74.3 \\
DARE-TIES & 20.0 & 29.2 &  92.4 &  0.8 & 85.3 &  72.9 \\
\bottomrule
\end{tabular}
\caption{\textbf{Balanced vs.\ full-data FT WJB.} CARES: 4-class accuracy;
Atk.: jailbreak refusal rate; Ben.: benign non-refusal rate. All values \%.}
\label{tab:datasize}
\end{table}

Three patterns hold. First, CARES accuracy under Linear, SLERP, and TIES stays at
or below chance, so the classification collapse is not explained by data scale
alone. Second, attack resistance falls throughout, as expected from a weaker
balanced anchor ($77.0\%$ vs.\ $88.0\%$). Third, DARE-TIES inverts---$20.0\%$ CARES
accuracy but only $29.2\%$ resistance---because with smaller WJB deltas its
stochastic pruning removes enough refusal-aligned updates for CARES signal to
survive.

\section{Attention Head Identification}
\label{app:heads}

\paragraph{Method}
Gemma-3-1B-IT uses multi-query attention with 26 layers and 4 query heads per layer for 104 total heads. For each model, we learn a binary necessity mask $m \in \{0,1\}^{26\times4}$ identifying the minimal head set that preserves task performance. Heads with $m=0$ are mean-ablated; their output-projection input is replaced by the activation mean over a 128-prompt calibration set. Masks are optimized by a $(1{+}\lambda)$ evolutionary strategy that minimizes retained heads subject to a $\pm3\%$ tolerance from the original baseline. We report one mask per model and treat exact head positions as one valid solution rather than canonical.

\begin{figure}[ht]
\centering
\includegraphics[width=\textwidth]{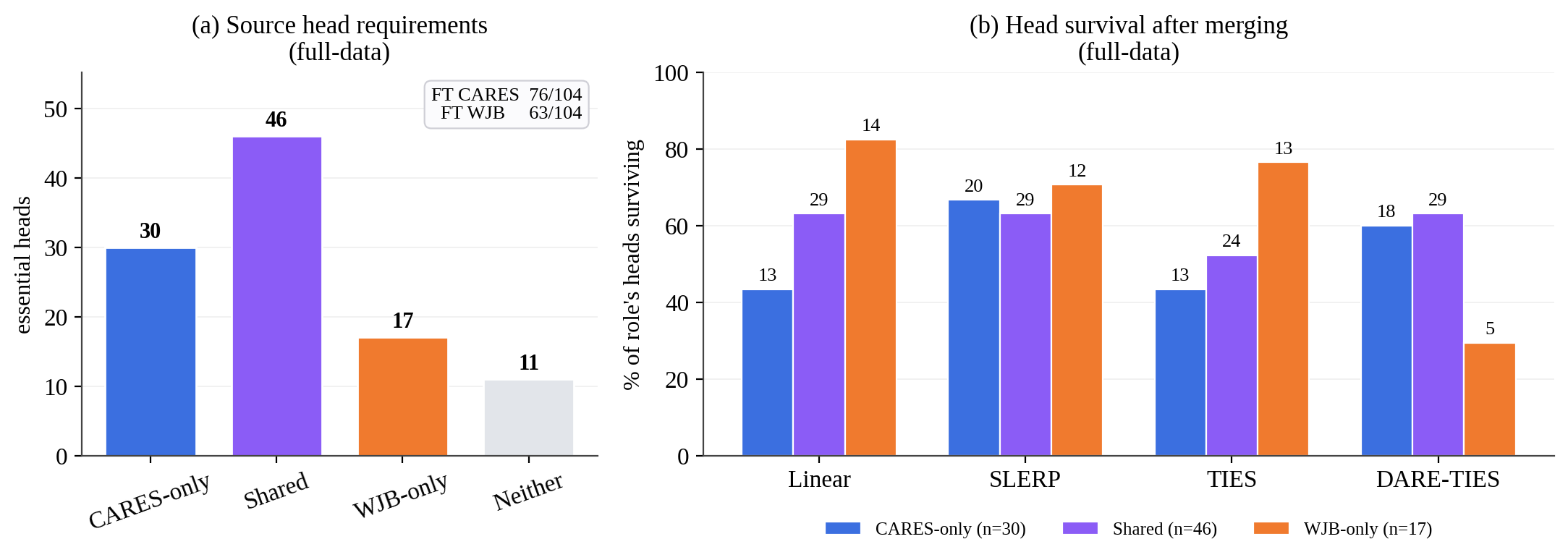}
\caption{\textbf{Attention-head necessity (full finetuned)} (a) Partition of the 104 heads by source role. (b) Survival of each role's heads after merging; bar height is the surviving fraction (role base $n$ in legend) and the number above each bar is the surviving head count. WJB-only heads survive at markedly higher rates than CARES-only heads under Linear and TIES.}
\label{fig:heads_full}
\end{figure}

\paragraph{Full Finetune Models}
FT CARES requires 76 of 104 heads; FT WJB requires 63. They share 46 (Jaccard $=0.49$), 30 CARES-only, 17 WJB-only, and 11 needed by neither (Fig.~\ref{fig:heads_full}a). Head survival after merging differs by method (Fig.~\ref{fig:heads_full}b). Linear and TIES preferentially destroy CARES-specific heads: only 13/30 CARES-only heads ($43\%$) survive under each, against 14/17 ($82\%$) WJB-only under Linear and 13/17 ($76\%$) under TIES. SLERP is roughly symmetric (20/30, $67\%$ vs.\ 12/17, $71\%$), and DARE-TIES reverses the pattern (18/30, $60\%$ CARES-only vs.\ 5/17, $29\%$ WJB-only), consistent with its stochastic sparsification concentrating refusal into a few heads. TIES is further destructive to shared heads (24/46, $52\%$, vs.\ 29/46, $63\%$ for the other methods), which further compounds CARES loss and matches its $0.6\%$ accuracy.
 
\begin{figure}[ht]
\centering
\includegraphics[width=\textwidth]{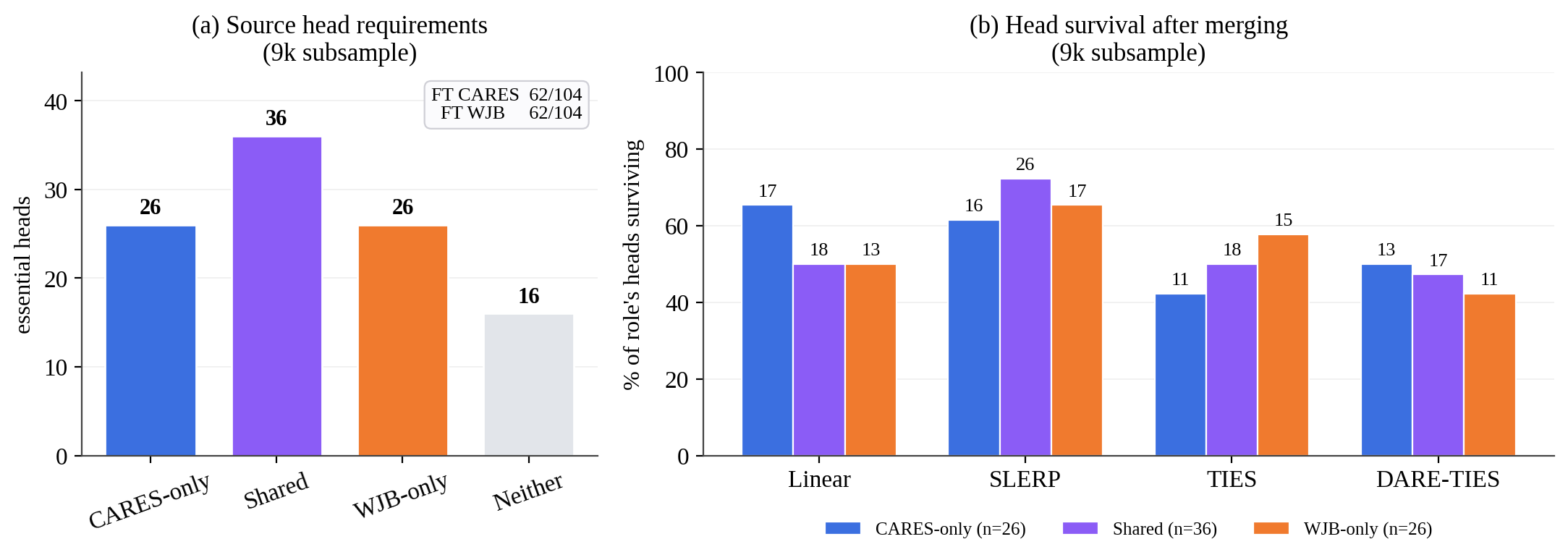}
\caption{\textbf{Attention-head necessity (9k subsample)} Same layout as Fig.~\ref{fig:heads_full}. With matched training data the CARES/WJB survival asymmetry is substantially reduced.}
\label{fig:heads_9k}
\end{figure}

\paragraph{9k subsample for WJB}
For the balanced data finetunes, sources require 62 heads (36 shared, Jaccard $=0.41$), partitioning into 26 CARES-only, 26 WJB-only, and 16 dispensable (Fig.~\ref{fig:heads_9k}a). The survival asymmetry largely disappears (Fig.~\ref{fig:heads_9k}b); both Linear and DARE-TIES retain more CARES only heads. Only TIES retains a WJB bias (15/26, $58\%$ vs.\ 11/26, $42\%$), most likely explained by its sign election amplifying even small magnitude gaps. This supports magnitude asymmetry as the cause of the selective destruction; symmetric retention does not, however, yield balanced capability, as both degrade.
 
\newpage

\begin{figure}[t]
\includegraphics[width=\textwidth]{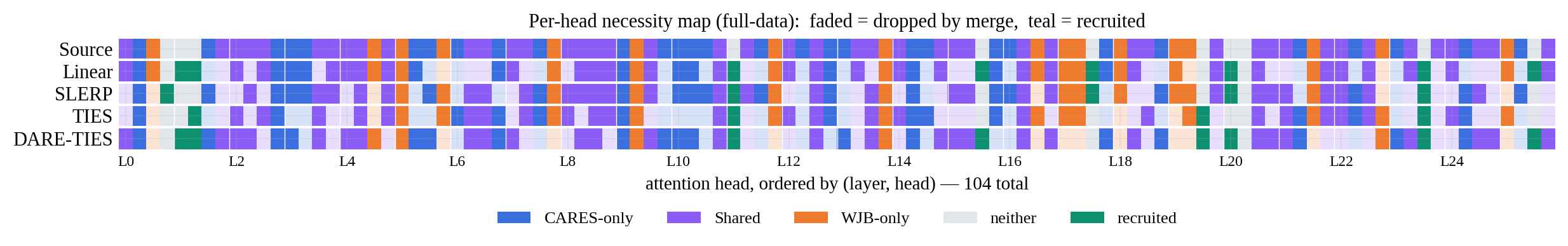}\\[3pt]
\includegraphics[width=\textwidth]{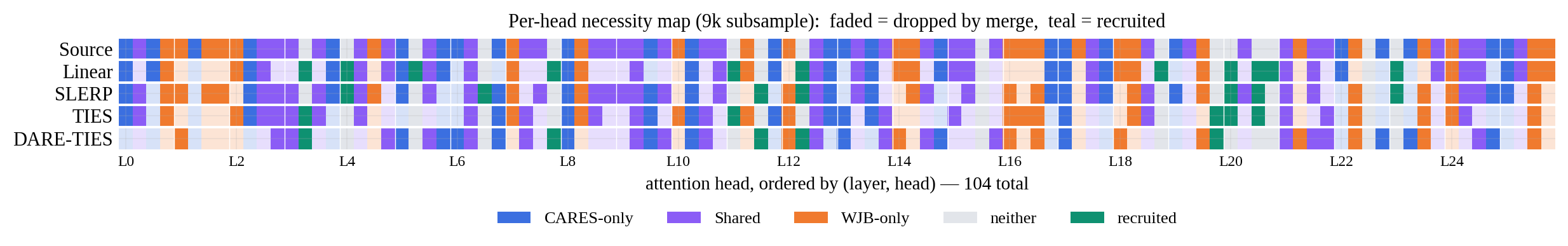}
\caption{\textbf{Per-head necessity map} Each cell is one attention head, ordered by (layer, head); the top row of each panel is the source-role assignment and the remaining rows are the four merges. Faded = head dropped by the merge; teal = head recruited from positions neither source required. \emph{Top:} full finetunes. \emph{Bottom:} 9k subsample. Heads lost under merging are spread across depth, and every method recruits a handful of new heads.}
\label{fig:head_map}
\end{figure}
 
\paragraph{Per-head map}
Figure~\ref{fig:head_map} resolves survival head-by-head rather than in aggregate. Each cell is one of the 104 heads, colored by source role; in the merge rows a faded cell marks a head dropped from the necessity mask and a teal cell marks a head recruited from positions neither source required. From these charts, we can first see that the loss is distributed rather than localized. Dropped CARES-only and shared heads are scattered across depth under every method, so merging degrades classification broadly rather than severing a single region. Second, merging is not purely subtractive, since each method recruits heads that neither source needed (8 under Linear and DARE-TIES, 5 under SLERP, 4 under TIES in the full-data run), indicating that weight averaging reorganizes computation, opening new dependencies as it overwrites old ones.

\end{document}